\documentclass{article}

\usepackage[preprint]{neurips_2026}

\usepackage[utf8]{inputenc} 
\usepackage[T1]{fontenc}    
\usepackage{hyperref}       
\usepackage{url}            
\usepackage{booktabs}       
\usepackage{amsfonts}       
\usepackage{nicefrac}       
\usepackage{microtype}      
\usepackage{xcolor}         
\usepackage{amsmath}
\usepackage{verbatim}
\usepackage{float}
\usepackage{graphicx}
\usepackage{amsmath,amssymb}
\usepackage{booktabs}
\usepackage{graphicx}
\usepackage{caption}
\usepackage{subcaption}
\usepackage{hyperref}
\usepackage{tabularx}
\usepackage{xcolor}
\usepackage{array}
\usepackage{longtable}

\newcommand{\rhoimp}{\ensuremath{\rho_{\text{imp}}}}
\newcommand{\rhodir}{\ensuremath{\rho_{\text{dir}}}}

\DeclareMathOperator{\Dec}{Dec}
\title{Evaluating the Causal Faithfulness of SHAP as a Candidate Attribution Method for SAE Features\thanks{Code is available at \url{https://github.com/froge79/shap-sae}}}

\author{%
  Samuel Xie\\
  Independent \\
  \And
  Noura Al-Moubayed \\
  Durham University \\
}

\begin{document}
\raggedbottom
\bibliographystyle{plainnat}

\maketitle

\begin{abstract}
  Sparse autoencoders decompose language model representations into interpretable features, but feature interpretation in this literature remains predominantly correlational. Attribution methods offer more direct estimates of feature influence, yet their faithfulness to actual causal effects has not been systematically evaluated in the SAE setting. We apply SHAP-based attribution to SAE latent features and evaluate it against direct causal intervention, using sentiment classification on SST-2 with GPT-2 Small and a 32,768-feature layer-7 SAE as a testbed. We compare SHAP to Integrated Gradients, Gradient$\times$Input, and probe-weight baselines across linear and MLP probe arms, scoring each against feature ablation and activation steering measured through the language model's own logit difference. Most faithfulness associations we measure do not survive multiple-comparison correction, and attribution methods that agree strongly with one another ($\rho = 0.42$--$0.80$) agree poorly with measured causal effects, indicating that inter-method agreement is not evidence of causal faithfulness. Two effects hold under control. First, local per-example attribution retains a directional advantage over global averaged attribution, while its apparent importance advantage is explained by feature activity rather than superior causal ranking. Second, under known-ground-truth minimal pairs, attribution recovers inserted sentiment signal well above chance (AUROC $0.792$) but fails to localize deletions (AUROC $0.520$), suggesting that these methods detect the presence of a feature's influence more reliably than its absence. We argue that importance faithfulness and directional faithfulness are separable properties requiring separate evaluation, and that attribution agreement should not substitute for causal validation.
\end{abstract}

\section{Introduction}
Sparse autoencoders (SAEs) have emerged as the leading tool for decomposing the internal representations of large language models into interpretable features. By training a sparse dictionary on residual stream activations, SAEs decompose superposed representations into nearly monosemantic directions that can be inspected individually \citep{bricken2023monosemanticity, huben2024sparse}. Recent work has scaled this approach to frontier models including Claude 3 Sonnet and GPT-4, recovering tens of thousands of interpretable features \citep{templeton2024scaling, gao2025scaling}.

Despite this progress, SAE feature interpretation remains largely correlational. The dominant pipeline identifies top-activating dataset examples for each feature and inspects them manually or with automated interpretability models to assign semantic labels—revealing the contexts in which a feature activates, but not establishing what inputs causally drive the feature or what downstream effects it causally produces. Attribution methods such as logit weight analysis, gradient-based attribution, and attribution patching provide more direct estimates of feature influence, but many rely on local linear approximations that may miss nonlinear interactions \citep{elhage2021mathematical, syed2024attribution}. To our knowledge, no prior work has systematically evaluated SHAP-based attribution of SAE features against direct causal interventions, nor examined whether any attribution methods' SAE feature rankings are faithful to measured causal effects.

To address the gap, we apply SHAP-based attribution to SAE latent features using sentiment classification on SST-2 as a testbed, evaluating multiple attribution methods---SHAP, Integrated Gradients, and $\text{Gradient}\times\text{Input}$---against causal interventions in the form of feature ablation and activation steering. This design allows us to ask not just which features attribution methods identify, but whether those identifications are causally faithful. 

\paragraph{Contribution.}
\begin{itemize}
    \item We provide the first systematic application of SHAP-based attribution to SAE latent features, comparing against Integrated Gradients, gradient-input, and probe weight baselines across linear and MLP probe architectures.
    \item We evaluate attribution faithfulness causally via feature ablation and activation steering, establishing intervention effect as ground truth rather than relying on correlational proxies.
    \item We demonstrate that local per-example SHAP attributions are more faithful to causal directional effects than global mean attributions, identifying averaging as a source of faithfulness degradation.
    \item We construct a known-ground-truth validation using minimal sentiment-word edits, finding that attribution methods recover feature-level causal effects above chance for insertion edits, but not for removal edits, suggesting attribution faithfulness is asymmetric with respect to the direction of intervention.
\end{itemize}

\section{Related Work}
\paragraph{Sparse Autoencoders for Interpretability.} Sparse autoencoders decompose residual stream activations into sparse, nearly monosemantic features, addressing the polysemanticity that makes individual neurons difficult to interpret \citep{bricken2023monosemanticity, huben2024sparse}. Recent work has scaled this approach to frontier models, recovering tens of thousands of interpretable features \citep{gao2025scaling}. Feature interpretation in this literature remains predominantly correlational, relying on top-activating examples or auto-interpretability pipelines rather than causal validation.

\paragraph{Attribution Methods.} Shapley-based attribution \citep{lundberg2017unified} and Integrated Gradients \citep{sundararajan2017axiomatic} are standard tools for identifying which input features drive model predictions. Attribution patching \citep{syed2024attribution} applies gradient-based attribution to transformer internals, providing a computationally efficient approximation to activation patching. No prior work has systematically applied these methods to SAE latent features or evaluated their causal faithfulness in that setting.

\paragraph{Faithfulness Evaluation.} Faithfulness measures the degree to which an explanation reflects a model's true decision-making process \citep{jacovi2020towards}. Prior evaluation typically relies on input perturbation metrics rather than direct causal intervention, conflating correlation with causation. Our work evaluates faithfulness via feature ablation and activation steering, using intervention effects as ground truth.

\paragraph{Probing.} Linear probes over internal representations are a standard tool for identifying whether a concept is encoded at a given layer \citep{alain2016understanding}. We use probes over SAE activations to define a prediction function over the latent feature space, through which attribution methods are applied.

\section{Methodology}
\subsection{Experimental Setup}
We evaluate on SST-2 \citep{socher2013recursive}, a binary sentiment classification dataset of single-sentence movie reviews. We partition the training split into three subsets with a fixed random seed. The train set is used for fitting probes, computing attribution backgrounds and baselines, and calibrating the steering parameter $\alpha$. The validation set is used for probe hyperparameter selection, constructing the activation-frequency feature mask, and reporting probe accuracy. The held-out set is reserved exclusively for all attribution scoring and causal evaluation (ablation and steering).

Evaluation and background rows are drawn from a fixed seeded permutation, ensuring nested samples (a 100-row sample exactly matches the first 100 rows of a 2000-row sample) and stable backgrounds across evaluation sizes. We use 2000 evaluation rows and 100 background rows throughout.

We experiment on frozen GPT-2 Small (124M parameters) \citep{radford2019language} with the \verb|gpt2-small-resid-post-v5-32k| SAE from \verb|sae-lens| \citep{gao2025scaling}, which decomposes layer 7 residual stream activations into 32,768 latent features. Activations are extracted at the last non-padding token position.

\subsection{Probing Models}
We train two probe architectures on SAE activations, each evaluated with identical attribution methods and causal evaluation.

A logistic regression classifier over the full 32,768-dimensional SAE activation vector, fit with L1 penalty ($C=1$, \verb|liblinear| solver). L1 regularization induces sparsity, yielding $\sim1400$ non-zero weights that define the filtered feature mask used in all downstream analyses. 

A one-hidden-layer MLP (width 32, ReLU, Adam, L2 regularization) fit on the filtered feature subset. The MLP is included because the linear probe's constant gradient collapses IG and GA to closed-form identities with SHAP, precluding substantive method comparison. We note a mild optimistic bias: the input mask was derived from the validation set, introducing a label-free but non-independent dependency.

\paragraph{Feature Mask.} Our feature mask combines two filters on the full val split only: a non-zero linear-probe coefficient and an activation frequency threshold of $5\%$. Our current mask results in 116 features, and only restricts the MLP probe's input layer, steering candidate pool, and group-steering ranking pool. It does not gate the linear probe's attribution methods.

\subsection{Attribution Methods}
Every method explains the same target scalar per arm: log-odds margin ($wx+b$) for the linear arm and pre-sigmoid logit for the MLP arm. Scales across arms are not directly comparable.

\paragraph{Zero-cost Baseline.}We report raw probe weight and saliency ($\|W_1[i,:]\|_2$) for the linear and MLP arms, respectively.

\paragraph{SHAP.}We use \verb|shap.LinearExplainer| on a deterministic 100-row background, computing exact Shapley values $\phi$ across all 32,768 features. For the MLP arm, we use \verb|shap.DeepExplainer| targeting pre-sigmoid logits over the 116 active input features, enforcing additivity against the background expectation.

\paragraph{Integrated Gradients \& Gradient $\times$ Input.}Integrated Gradients (IG) is evaluated via a 50-step trapezoidal sum for the MLP arm, whereas in the linear arm it reduces to closed-form $(x_j - \text{baseline}_j) w_j$. Single-point Gradient $\times$ Input ($\text{GA}_j = (x_j - \text{baseline}_j) \frac{\partial f}{\partial x_j}(x)$) equals linear IG by constant gradients and is included for arm structural parity.

\subsection{Feature-Ranking Agreement}
We compute pairwise Spearman $\rho$ between attribution methods, restricting evaluation to features with non-zero probe scores to prevent artificial inflation from zero-weighted features. For the linear probe, IG, GA, and SHAP are analytically identical, serving as a sanity check ($\rho=1$). MLP agreement is evaluated on $|\text{score}|$ since raw probe saliency lacks directional sign.

\subsection{Causal Evaluation}
Attribution methods (SHAP,IG,GA) are correlational and describe a fixed model with no intervention. To test causal faithfulness, we intervene on SAE feature activations and measure the resulting change in model behavior via the model's own logit difference. We read activations from a hook attached to the residual stream output of \verb|gpt-2 small| layer 7, the same layer used for probe training and attribution computation. For feature $i$ with activation $a_i$:

\textbf{Ablation:}
\[
\tilde{a}_i = 0, \quad \tilde{z} = \Dec(\tilde{\mathbf{a}})
\]
\noindent\textbf{Steering (additive, signed):}
\[
\tilde{a}_i = a_i + \alpha_i, \quad \tilde{z} = \Dec(\tilde{\mathbf{a}})
\]
 We pass the residual stream activations into the SAE, then encode and decode with no edit as a baseline to isolate the SAE reconstruction error from the actual intervention effects. We apply the intervention at every token position. However, to partially address the scope mismatch between last-token attribution and all-token intervention, we additionally evaluate a last-token-only steering condition in which the intervention is applied exclusively at the final token position, matching the attribution scope; results for last-token steering is reported in Table~\ref{tab:all-selection-faithfulness-last}, and all-token steering is reported in Appendix~\ref{app:all-token-global-faithfulness}.

The causal readout is the logit difference at the final token:
 \begin{equation*}
     \Delta = \ell(\text{wonderful''}) - \ell(\text{awful''}),
 \end{equation*}
 evaluated before and after the intervention. We report both $|\Delta|$ (importance) and signed $\Delta$ (direction). Because ablation actually reflects a negative $\Delta$, we negative raw $\Delta$ before correlation, so positive $\rho$ always indicates more faithfulness. 

\label{para:steering_calibration}We calibrate the global steering multiplier $\alpha$ on the training split using a sweep across $\{0.5,1,2,4\} \times \text{median natural positive activation scale}$ over three top-$|\text{SHAP}|$ features across $200$ examples. We select the smallest $\alpha$ achieving a mean probability shift $|\Delta P| \ge 0.02$ on a Layer 11 probe while keeping output saturation ($P < 0.02$ or $P > 0.98$) below $25\%$. We scale steering per feature ($\alpha_i = \alpha \cdot \frac{\text{scale}_i}{\text{median}(\text{scale})}$) to normalize nudges across varying baseline activation magnitudes, running unscaled $\alpha$ as a robustness check. Decoder weights are unadjusted as they are already near unit-norm ($\text{median} = 0.9996$).

We evaluate three candidate selection strategies: (1) uniform random sampling of $k$ features, (2) top-$k$ selection by $|\text{SHAP}|$ (noting this range-restricts SHAP's scored variance and artificially deflates its $\rho$), and (3) full evaluation over all 116 filtered features without resampling. Additionally, to control for random sampling noise, we evaluate random draws across 10 random seeds. For each method and correlation type, we test for significant non-zero correlation using a one-sample Wilcoxon signed-rank test alongside a 95\% bootstrapped confidence interval and an exchangeability check for seed draw typicality.

\subsection{Local Faithfulness}\label{subsection:local_faithfulness}
Global attribution $\Phi$ averages across examples, which can obscure per-example attribution quality or mask local disagreement. For each sentence $x$, we compute local SHAP $\phi(x)$. To ensure equal representation across methods, candidate features for each $x$ are selected via $\text{top-}k \text{ by } |\phi(x)| \cup \text{top-}k \text{ by } |\Phi|$. Intervention effects are tracked per-example, and we measure Spearman $\rho\big(|\text{local } \phi(x)|, |\Delta(x,\cdot)|\big)$ against $\rho\big(|\text{global } \Phi|, |\Delta(x,\cdot)|\big)$ over identical candidate sets. We report mean $\rho$ across sentences ($\pm\text{std}$) and the win-rate fraction where local outperforms global.

\paragraph{Confound.}\label{para:confound}Globally-top features ($|\Phi|$) are frequently inactive in a given single sentence. Because ablating an inactive feature moves the logit by zero, global attribution's lower correlation is partly an artifact of predicting non-zero effects for inactive features, rather than proof of superior ranking quality by local attribution. To isolate this confound, we evaluate under two regimes: Block [A], which includes all candidate features, and Block [B], which restricts candidates exclusively to active features ($\text{activation} > 0$).
\paragraph{Paired significance tests.} We conduct Wilcoxon signed-rank tests for $\rho_\text{local}-\rho_\text{global}$ with a bootstrap CI on the mean gap for effect size. We run for both blocks and both correlation types. 

We run two follow-up checks:
\begin{enumerate}
    \item \textbf{Decomposition of the block [A] margin} regresses the per-example gap on the active fraction of candidates using two variations (a simple linear regression and a size-controlled variant) to quantify the confound.
    \item \textbf{Continuous activity control} uses a partial Spearman of $|\text{attribution}|$ vs $|\Delta|$ controlling for $|\text{attribution}|$ to control for the possibility that local SHAP simply ranks features by activation strength. 
\end{enumerate}

\subsection{Multi-Feature Steering}
As a secondary analysis, we constructed steering vectors by aggregating the decoder directions of the top-\(k\) features selected by each attribution method and evaluated their effects on model outputs across intervention strengths. Full experimental details and results are provided in Appendix~\ref{app:group-full}.

\subsection{Known-Ground-Truth Validation}
Natural SST-2 text lacks ground truth feature importance, leaving near-zero faithfulness $\rho$ ambiguous between attribution failure and harness insensitivity. We construct minimal pairs with known causal sources of sentiment change: primary stimuli delete the single sentiment lexicon word from a sentence, while secondary stimuli append a sentiment clause to a neutral sentence. Expected direction is known in both cases.

We measure three quantities in dependency order: (1) \textbf{readout validity} (whether logit-difference sign tracks known edit direction, a precondition for faithfulness claims); (2) \textbf{feature recovery} (AUROC of each attribution method at identifying features whose activation changed, against a $0.5$ chance baseline); and (3) \textbf{causal recovery} (correlation between per-feature ablation effects and attribution scores on the edited sentence).

We explicitly control for two confounds: setting the SHAP background to the pair's original activations (which tautologically inflates AUROC by making scores proportional to activation change) and selecting distractors by top global SHAP magnitude (which forces its AUROC toward zero). We correct both by using the canonical training background for SHAP and sampling distractors randomly.

\subsection{Lexical Simplicity Control}
We evaluate whether SST-2 is sufficiently lexically simple that interaction-aware attribution cannot outperform simpler methods, using the full 10,103-example held-out split. 

We conduct three sub-analyses: (1) \textbf{co-activation} (mean active features per example and pairwise Jaccard similarity among top-20 features to test sparsity); (2) \textbf{lexical tracking} (point-biserial correlation between top feature activations and sentiment lexicon words to test if features act as simple word detectors); and (3) \textbf{lexicon baseline} (held-out accuracy of a bag-of-words logistic regression versus the SAE probe to assess value added beyond word-spotting).

\section{Results}
\subsection{Probe Performance}
Table~\ref{tab:probe-performance} shows the mean probe accuracy and AUROC for both the linear and MLP probe. Both probes exhibit substantial above-chance performance, while the MLP probe's validation remains mildly optimistic due to pre-filtered input features from the feature mask. 
\begin{table}[h!]
\centering
\caption{Probe architectures and performance.}
\label{tab:probe-performance}
\begin{tabular}{lccc}
\toprule
Arm & Input dim & Val.\ Accuracy & Val.\ AUROC \\
\midrule
Linear (logistic) & 32{,}768 & 0.761 & 0.84 \\
MLP (hidden=32)   & 116 (filtered) & 0.735 & 0.808 \\
\bottomrule
\end{tabular}
\end{table}

\subsection{Attribution Method Agreement}
\begin{figure}[H]
\centering
\includegraphics[width=0.6\linewidth]{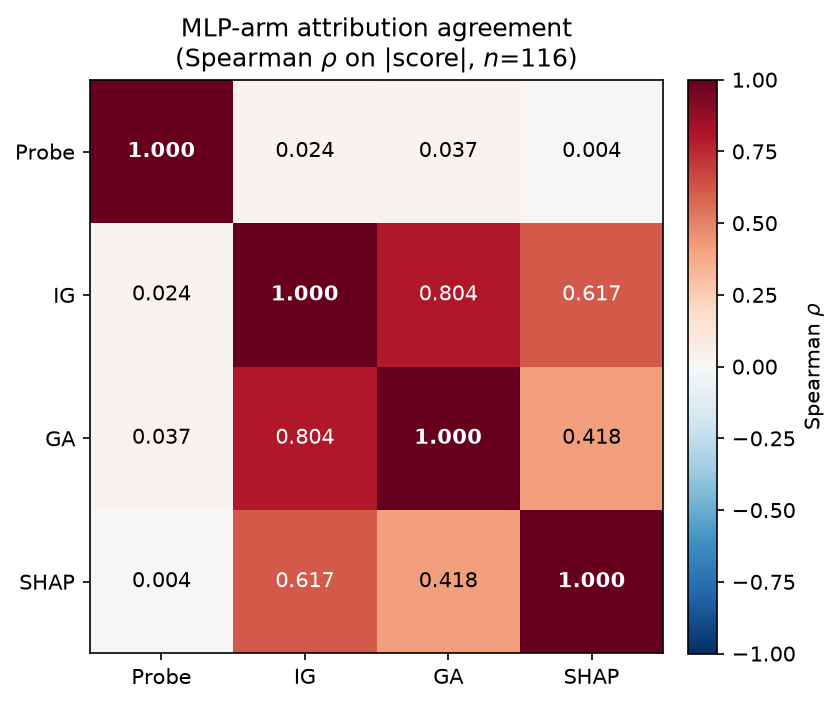}
\caption{MLP-arm attribution agreement (Spearman $\rho$ on $|\text{score}|$, $n=116$ filtered features). IG, GA, and SHAP form a correlated cluster ($\rho=0.42$–$0.80$); the probe weight is essentially uncorrelated with all three ($\rho\leq0.04$).}
\label{fig:mlp-ranking-agreement}
\end{figure}

As illustrated in Figure~\ref{fig:mlp-ranking-agreement} and Table~\ref{tab:app-mlp-ranking}, the gradient-based methods (IG and GA) correlate strongly with each other and moderately with SHAP, while the probe correlates with none of them. The probe weights were largely uncorrelated with the gradient and SHAP-based attributions, whereas IG, GA, and SHAP showed moderate-to-strong agreement.

\subsection{Causal Faithfulness}

Table~\ref{tab:all-selection-faithfulness-last} reports the powered global faithfulness evaluation across all 116 filtered features under last-token steering, BH-corrected within arm. In the linear arm, SHAP/IG/GA achieve moderate importance faithfulness ($\rho_{\text{imp}} = 0.256$, $q = 0.0111$), while probe weights show strong directional faithfulness ($\rho_{\text{dir}} = 0.410$, $q < 0.0001$); SHAP/IG/GA directional faithfulness is near zero and does not survive correction. In the MLP arm, SHAP and IG achieve importance faithfulness above the FDR threshold ($\rho_{\text{imp}} = 0.279$ and $0.278$, both $q = 0.0088$), while directional faithfulness remains nonsignificant across all MLP methods. Overall, 4/8 linear-arm tests and 2/7 MLP-arm tests survive FDR correction, with importance faithfulness more consistently detected than directional faithfulness across both arms.
\begin{table}[H]
    \centering
    \caption{Powered global faithfulness replication ($k=116$, steering, all filtered features, BH-corrected within arm).}
    \label{tab:all-selection-faithfulness-last}
    \begin{tabular}{llcc}
    \toprule
    Arm & Method & $\rho_\text{imp}$ ($q$) & $\rho_\text{dir}$ ($q$) \\
    \midrule
    Linear & SHAP/IG/GA & $+0.256$ ($q$=0.0111)* & $-0.054$ ($q$=0.563) \\
    Linear & Probe      & $+0.096$ ($q$=0.487)  & $+0.410$ ($q$<0.0001)* \\
    \midrule
    MLP & SHAP           & $+0.279$ ($q$=0.0088)* & $-0.157$ ($q$=0.162) \\
    MLP & Probe saliency & $-0.011$ ($q$=0.911) & n/a \\
    MLP & IG             & $+0.278$ ($q$=0.0088)* & $-0.046$ ($q$=0.862) \\
    MLP & GA             & $+0.197$ ($q$=0.0797) & $-0.031$ ($q$=0.862) \\
    \bottomrule
    \multicolumn{4}{l}{\footnotesize *survives FDR 0.05. Linear arm: 4/8 tests survive; MLP arm: 2/7.}
    \end{tabular}
    \end{table}
Additionally, last-token steering yields higher importance faithfulness than all-token steering in both arms and recovers MLP-arm significance that the all-token condition did not (Appendix~\ref{app:all-token-global-faithfulness}), suggesting that the scope mismatch attenuates measured faithfulness when the intervention extends beyond the attributed token position.

\subsection{Local Faithfulness}
\begin{figure}[H]
\centering
\includegraphics[width=\linewidth]{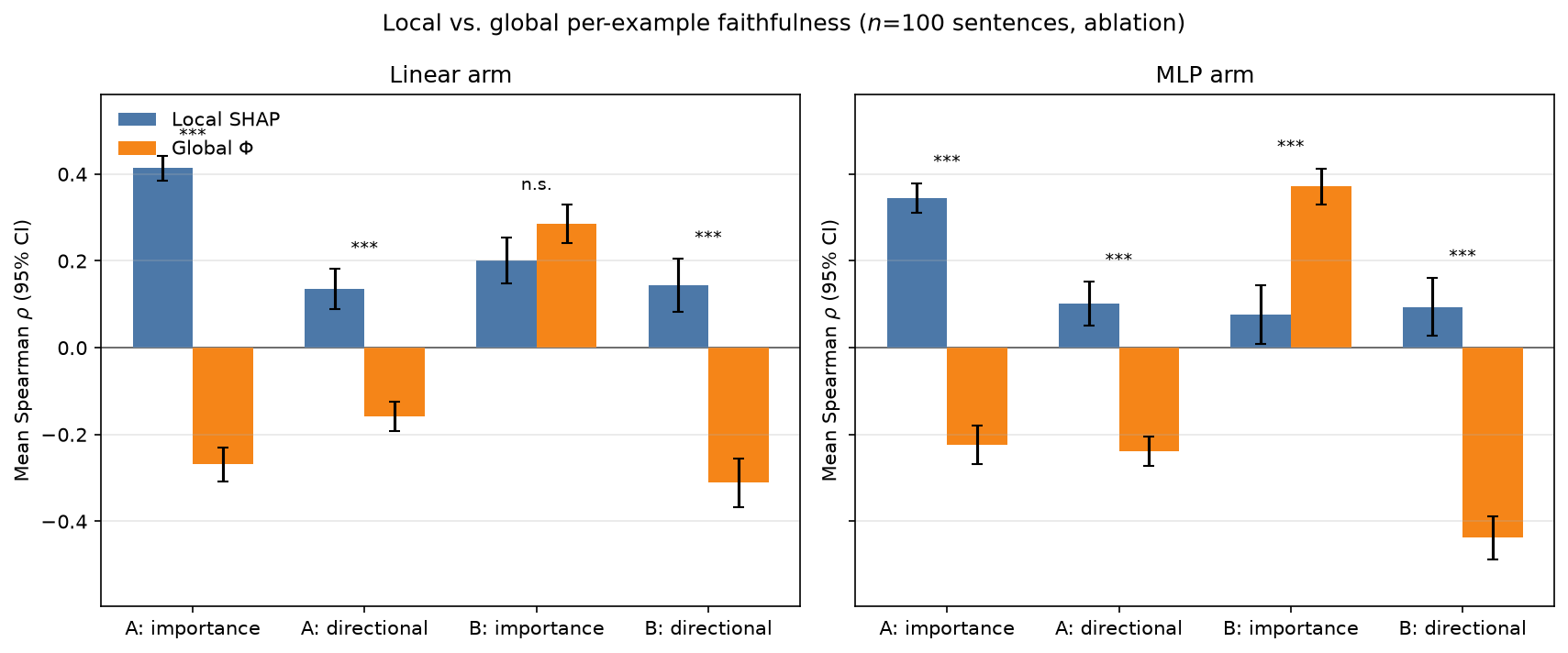}
\caption{Local vs.\ global attribution, per-example faithfulness ($n=100$ sentences, ablation). Bars are mean Spearman $\rho$ across examples $\pm$ 95\% CI; asterisks mark the paired Wilcoxon signed-rank test on the per-example gap (*** $p<10^{-4}$, ** $p<10^{-2}$, * $p<0.05$, n.s.\ otherwise). Block A = all top-20-local $\cup$ top-20-global candidates; Block B = active-only control.}
\label{fig:local-vs-global}
\end{figure}

Figure~\ref{fig:local-vs-global} outlines local vs. global SHAP causal faithfulness results. Importance fails to survive the activity control in the both arms, while directional advantage persists across both arms, suggesting that local attribution carries per-example directional information that averaging into $\Phi$ destroys. Overall, local SHAP improves directional faithfulness, but its apparent advantage in importance faithfulness is explained by activity information rather than superior causal ranking among active features.

Our full local vs. global attribution results are shown in Table~\ref{tab:app-local-vs-global}. Local attribution substantially outperformed global attribution in Block A for both importance and directional faithfulness. However, activity explained only a modest fraction of the between-example variation in the local–global gap (\(R^2=0.019\) linear; \(0.081\) MLP; \(0.120/0.215\) controlling for candidate-set size). In Block B, restricting to active features eliminated the local advantage for importance, while its directional advantage persisted in both arms.

\begin{table}[H]
\centering
\caption{Local vs.\ global attribution, per-example faithfulness ($n=100$ sentences, ablation): numeric detail. Mean $\rho$ across examples with a paired Wilcoxon signed-rank test on the per-example gap.}
\label{tab:app-local-vs-global}
\begin{tabular}{llccccc}
\toprule
Arm & Block : kind & Local & Global & Gap & Wilcoxon $p$ & Win frac.\\
\midrule
Linear & A : importance  & $+0.414$ & $-0.269$ & $+0.682$ & $3.9\times10^{-18}$ & 1.000 \\
Linear & A : directional & $+0.135$ & $-0.158$ & $+0.294$ & $4.1\times10^{-15}$ & 0.860 \\
Linear & B : importance  & $+0.202$ & $+0.285$ & $-0.083$ & $6.6\times10^{-2}$  & 0.420 \\
Linear & B : directional & $+0.144$ & $-0.311$ & $+0.455$ & $9.1\times10^{-16}$ & 0.880 \\
\midrule
MLP & A : importance  & $+0.345$ & $-0.223$ & $+0.568$ & $3.9\times10^{-18}$ & 1.000 \\
MLP & A : directional & $+0.102$ & $-0.239$ & $+0.341$ & $1.2\times10^{-15}$ & 0.880 \\
MLP & B : importance  & $+0.077$ & $+0.371$ & $-0.295$ & $9.6\times10^{-9}$  & 0.270 \\
MLP & B : directional & $+0.094$ & $-0.438$ & $+0.532$ & $5.7\times10^{-17}$ & 0.920 \\
\bottomrule
\end{tabular}
\end{table}

\subsection{Known Ground-Truth Validation and Lexical Simplicity Control}
We evaluate attribution faithfulness under known-ground-truth conditions using 100 minimal pairs across two edit types---results are shown in Table~\ref{tab:minimal-pairs}. Readout validity is moderate in both conditions (sign accuracy 0.770 for remove, 0.700 for insert), establishing a sufficient but imperfect behavioral signal. For the remove condition, all three attribution methods recover edited features at near-chance levels (AUROC: local SHAP 0.520, global SHAP/IG/GA 0.509, probe weights 0.503), and causal recovery is weak across the board ($\rho=0.254$, 0.163, $-0.099$ respectively). The insert condition yields substantially stronger results: local SHAP and global SHAP/IG/GA achieve AUROC of 0.792, with probe weights lagging at 0.610; causal recovery follows the same ordering ($\rho=0.536$, 0.458, 0.167).
\begin{table}[H]
\centering
\caption{Minimal-pair ground-truth validation, $n=100$ pairs, random distractors.}
\label{tab:minimal-pairs}
\begin{tabular}{lccccc}
\toprule
Edit & Sign acc.\ & Method & AUROC & $\rho$ (causal recovery) \\
\midrule
remove & 0.770 & local SHAP  & 0.520 & 0.254 \\
       &       & global SHAP/IG/GA & 0.509 & 0.163 \\
       &       & Probe weights & 0.503 & $-0.099$ \\
\midrule
insert & 0.700 & local SHAP  & 0.792 & 0.536 \\
       &       & global SHAP/IG/GA & 0.792 & 0.458 \\
       &       & Probe weights & 0.610 & 0.167 \\
\bottomrule
\end{tabular}
\end{table}

Our lexical baseline provides substantially weaker held-out performance than the SAE probe (accuracy 0.629 vs. 0.761), while top-feature activation showed only weak association with sentiment-lexicon presence (mean $|r|=0.063$; Appendix~\ref{app:lexical}), suggesting that the probe signal is not reducible to simple sentiment-word detection.

\section{Discussion}

We apply SHAP to SAE feature attribution and evaluate it against a causal harness rather than agreement with other attribution methods. The principal outcome is negative: most faithfulness associations do not survive multiple-comparison correction. Two of four linear-arm tests clear FDR, two MLP-arm tests survive, and none of the \(k=20\) associations survive pooled correction across the multi-seed sweep. The \(k=116\) evaluation scores the entire filtered pool without resampling and therefore avoids this seed sensitivity, but the broader pattern indicates that a single \(k=20\) draw is insufficient for a faithfulness claim.

The disagreement between attribution rankings and measured causal effects is our central observation. IG, GA, and SHAP form a correlated cluster in the MLP arm (\(\rho=0.42\)--\(0.80\)) while corresponding poorly with intervention effects, indicating that inter-method agreement should not be treated as evidence that a method identifies causally important SAE features. This is not attributable to lexical simplicity: activation of top features is only weakly associated with sentiment-lexicon presence (mean \(|r|=0.063\)), and the SAE probe exceeds a bag-of-words baseline by 13.2 accuracy points.

The local-versus-global analysis separates two properties commonly conflated under faithfulness. Local SHAP's advantage in importance faithfulness does not survive activity control; in the MLP arm it reverses, with global attribution outperforming local (\(+0.371\) vs. \(+0.077\), \(p=9.6\times10^{-9}\)), suggesting that averaging into \(\Phi\) preserves information about which features are consistently causally relevant. Directional faithfulness behaves oppositely, with local attribution retaining its advantage in both arms under active-only and continuous activation controls. Importance and directional faithfulness are therefore separable properties, making the choice between local and global attribution criterion-dependent rather than uniformly advantageous.

The asymmetry between edit conditions carries a practical implication. Attribution recovers inserted sentiment signal well above chance (AUROC \(0.792\)) but fails to localize deletions (AUROC \(0.520\) and \(0.509\)), despite a readout registering the deletion correctly in 77\% of pairs. Sign accuracy exceeding feature recovery suggests that deletion redistributes activation across many features, each moving too little to detect. Attribution should therefore be expected to identify the presence of a feature's influence more reliably than its absence, and a null ablation result should not be read as evidence that no feature carries the effect.

These results are obtained for a single model, SAE, layer, and task. Whether the patterns generalize remains open; the decomposition-and-control methodology is not specific to SHAP and extends directly to other settings.

\subsection{Limitations}
\begin{itemize}
    \item Attributions are computed at a single token position. While restricting interventions to the last token improves importance faithfulness for non-linear attribution methods, it does not resolve the broader failure of directional faithfulness.
    \item The steering magnitude was calibrated against one readout but evaluated against another, a mismatch shown by the sign-inversion investigation to be non-trivial in small samples. 
    \item All faithfulness correlations are computed over a small number of features and remain underpowered to rule out modest true effects even where correction has been applied.
    \item Although the powered seed-0 analysis identifies several linear-arm associations, none survive pooled FDR correction across the multi-seed robustness sweep, indicating sensitivity to candidate sampling.    
\end{itemize}

\subsection{Future Directions}
The primary open question is whether the directional advantage of local attribution generalizes beyond sentiment classification and this specific SAE; the underlying mechanism is not specific to SHAP and the decomposition-and-control methodology developed here could extend to other settings. 

\section{Conclusion}
We evaluated SHAP as an attribution method for SAE latent features against direct causal intervention, comparing it to Integrated Gradients, Gradient$\times$Input, and probe-weight baselines across linear and MLP probe arms. Most measured faithfulness associations did not survive correction, and agreement among attribution methods did not track agreement with causal effects. Two findings held under control: local attribution retains a directional advantage that averaging destroys, while its apparent importance advantage reflects feature activity rather than superior causal ranking; and attribution recovers inserted sentiment signal well above chance but fails to localize deletions. We take these results to indicate that attribution agreement is not a substitute for causal validation, and that importance and directional faithfulness warrant separate evaluation.

\section{Use of AI Tools}
The authors used Claude Sonnet 5 and Opus 5 to assist with generating portions of the experimental code and with drafting and editing the paper’s prose. The authors reviewed, tested, and verified all AI-assisted code and text, and take full responsibility for the experiments, results, and conclusions reported here.

\bibliography{references}

\clearpage
\appendix

\section{Experimental Setup Details}
\label{app:setup}

In this appendix, we give implementation details underlying the main text's
results: the dataset and splits, the model and sparse autoencoder, the
probe architectures, and the feature-selection procedure, focusing on
choices that bear directly on the validity of the reported numbers.

\subsection{Task and Dataset}
\label{app:dataset}

We study binary sentiment classification on SST-2 (Stanford Sentiment
Treebank, binary variant), loaded via the HuggingFace
\texttt{stanfordnlp/sst2} release. We split the working dataset once, with
a fixed seed (seed$=42$), into three disjoint subsets, which we never
regenerate:

\begin{table}[h]
\centering
\caption{Data split sizes and purposes.}
\label{tab:app-splits}
\begin{tabularx}{\textwidth}{l r X}
\toprule
Split & $n$ & Used for \\
\midrule
Probe train & 47{,}144 & Fitting probes; SHAP/IG/GA background and baseline; steering $\alpha$ calibration \\
Probe val & 10{,}102 & Probe hyperparameters; the activation-frequency feature mask; reported probe accuracy \\
Held-out (``shap'') & 10{,}103 & All attribution (SHAP/IG/GA) and all causal evaluation (ablation/steering) \\
\bottomrule
\end{tabularx}
\end{table}

This split is the central methodological invariant of the study: we
compute both attribution and causal evaluation on the held-out split, on
which we fit or tune nothing, so a faithfulness comparison between an
attribution method's prediction and an intervention's measured effect is
not a fit-the-training-data tautology.

\paragraph{Canonical row sampling.} We draw all evaluation and background
rows through a single function (seed$=42$) that returns \emph{prefixes} of
one seeded permutation, not independent draws. Two properties follow: (i)
the 100-row local sample is exactly the first 100 rows of the 2000-row
global sample, so local and global results join by row position; and (ii)
background sampling from the train split (n$=100$) is unaffected by the
evaluation sample size. We use canonical sample sizes of
$n_{\text{eval}}=2000$ (held-out) and $n_{\text{background}}=100$ (train)
throughout; individual experiments may use a prefix of this (e.g.\ 1000 for
the headline steering runs, 100 for local per-example runs).

\subsection{Model and Token Position}
\label{app:model}

We use GPT-2 Small and a 32{,}768-feature SAE at layer 7
(\texttt{resid\_post}) frozen throughout. Two implementation choices here
are non-obvious and bear directly on downstream validity:

\begin{itemize}
  \item \textbf{Token position}: we take the sentence-level representation
  to be the SAE activation at the \emph{last real (non-padding) token},
  resolved via a single canonical function that correctly handles GPT-2's
  pad\,=\,bos\,=\,eos\,=\,\texttt{<|endoftext|>} collision. We share this
  definition across extraction, probing, and every causal readout in the
  pipeline.
  \item \textbf{Decoder geometry}: SAE decoder ($W_{\text{dec}}$) rows are
  approximately unit norm (median 0.9996 across all 32{,}768 features); we
  therefore apply no decoder-norm correction to any steering or
  group-steering vector (Appendices~\ref{app:causal} and~\ref{app:group}).
\end{itemize}

\subsection{Probing Models}
\label{app:probes}

We train two probe architectures on SAE activations, forming two parallel
``arms'' of the study; we duplicate every attribution method and
causal-evaluation script across both.

\paragraph{Linear (logistic) probe --- the linear arm.}
We fit an L1-penalized \texttt{sklearn} logistic regression over the
full-width, 32{,}768-d SAE activation at the last token. The L1 penalty
zeroes the large majority of coefficients ($\sim$1.4k--1.3k of 32{,}768
non-zero, run-dependent), and we use this sparsity pattern as one of two
filters defining the ``filtered feature set'' used throughout
(Appendix~\ref{app:mask}). Seeding is load-bearing: an unseeded fit
re-rolls which coefficients land at exactly zero, which re-rolls the
downstream feature mask, the MLP probe's inputs, and the steering
candidate pool; we therefore fix \texttt{random\_state} on every
\texttt{LogisticRegression} fit in the pipeline (probe, residual probe,
lexicon baseline).

\paragraph{MLP probe --- the MLP arm.} We train a one-hidden-layer
\texttt{sklearn} MLP classifier on the \emph{filtered} feature subset only
(Appendix~\ref{app:mask}), not the full 32{,}768-d activation. We include a
second arm because the linear probe's gradient is constant, which makes
Integrated Gradients and Gradient$\times$Input collapse to a closed-form
identity with SHAP (Appendix~\ref{app:identity}) --- a sanity check rather
than a comparison. The MLP's ReLU nonlinearity makes the gradient
input-dependent, so IG, GA, and SHAP can genuinely disagree; this is where
the substantive method comparison lives.

The MLP probe's validation accuracy is mildly optimistic: we pre-select its
input features with a val-derived (label-free) activation-frequency
filter, so there is a small but real leakage between feature selection and
reported accuracy.

\subsection{Filtered Feature Set}
\label{app:mask}

We compute the ``feature mask'' consumed everywhere downstream with a
single function, combining two filters computed on the \emph{validation
split only} (never held-out, to avoid letting the evaluation split shape a
model that we later score on it):

\begin{enumerate}
  \item Non-zero linear-probe coefficient.
  \item Activation frequency $\geq 5\%$ on the validation split.
\end{enumerate}

This mask contains 116 features (of 32{,}768) in the experimental run we
report here. This count is not a fixed architectural constant: we refit it
whenever we refit the linear probe, and it depends on the probe's
coefficient sparsity pattern (Appendix~\ref{app:probes}). The mask gates
the MLP probe's input layer, the steering candidate pool (both arms), and
the group-steering ranking pool. It does \emph{not} gate the linear arm's
own SHAP/IG/GA computation, which we carry out over the full
32{,}768-width space and only slice to the mask where a downstream
consumer needs it.

\clearpage
\section{Attribution Method Definitions}
\label{app:attribution}

We compute three attribution methods per arm --- SHAP, Integrated
Gradients (IG), and Gradient$\times$(Input$-$Baseline) (GA) --- and report
raw probe weight/saliency as a zero-cost baseline.

\subsection{Target-Function Matching}
\label{app:target-matching}

We require every method within one arm to explain the \emph{same} scalar
function of the input; otherwise the cross-method comparison is confounded
by disagreement about what is being attributed rather than about
attribution quality. 

\begin{table}[h]
\centering
\caption{Target function attributed by each arm.}
\label{tab:app-target}
\begin{tabularx}{\textwidth}{l l X X}
\toprule
Arm & Explainer & Target function & Gradient used by IG/GA \\
\midrule
Linear (logistic) & \texttt{shap.LinearExplainer} (exact) & Log-odds margin, $w \cdot x + b$ & \texttt{coef\_} (constant) \\
MLP & \texttt{shap.DeepExplainer} & Pre-sigmoid logit & Input-dependent, through ReLU \\
\bottomrule
\end{tabularx}
\end{table}

We do not treat magnitudes as comparable across arms: a log-odds margin and
a pre-sigmoid logit are different scales, so only within-arm rankings are
comparable.

\subsection{SHAP --- Linear Arm}
\label{app:shap-linear}

We use \texttt{shap.LinearExplainer}, exact for a linear model (no sampling
approximation), with an \texttt{Independent} masker over a 100-row train
background; we pin \texttt{max\_samples} to the full background count to
avoid the library silently subsampling an unseeded random 100 rows from a
larger background. This explains all 32{,}768 features (cheap and exact, so
there is no reason to restrict it). We report
$\Phi_j = \mathbb{E}_e[\phi_j(x_e)]$, the \emph{mean signed} SHAP value per
feature over the 2000-row canonical held-out sample --- we preserve sign,
and it is the basis for every directional/causal comparison in
Appendix~\ref{app:causal}. We report per-feature sign consistency (the
fraction of examples whose SHAP value shares the majority sign) as a
secondary diagnostic of how representative $\Phi_j$ is of individual
examples.

\subsection{SHAP --- MLP Arm}
\label{app:shap-mlp}

We run \texttt{shap.DeepExplainer} on a hand-cloned PyTorch
\texttt{nn.Sequential(Linear, ReLU, Linear)} that reproduces the fitted
\texttt{MLPClassifier} exactly (we copy the weights in; a custom
\texttt{forward} would break DeepSHAP's op-level attribution). We leave
\texttt{check\_additivity=True} enabled and treat a failure there as a
real bug, not numerical noise to suppress. We target the pre-sigmoid logit
(DeepSHAP is documented as more reliable on logits than on post-sigmoid
probabilities), and restrict scope to the filtered feature set only,
matching the MLP's native input space.

\subsection{Integrated Gradients}
\label{app:ig}

We define IG generally as:
\[
\mathrm{IG}_j = (x_j - b_j) \cdot \int_0^1 \frac{\partial f}{\partial x_j}\big(b + \alpha(x-b)\big)\, d\alpha,
\]
and approximate this in the MLP arm with a Riemann sum over the path. In
the linear arm, the path integral of a constant gradient collapses in
closed form to $\mathrm{IG}_j = (x_j - b_j) \cdot w_j$, which we compute
directly rather than by numerically integrating a constant. In the MLP arm
the gradient is genuinely input-dependent (propagated through the ReLU
activation mask), so the Riemann sum performs real numerical integration.
We take the baseline $b$ to be the \emph{mean} of the 100-row train
background, not a zero vector --- this matters because SAE features are
$\sim$87\% zero, and a zero baseline would make GA (Appendix~\ref{app:ga})
degenerate to 0 whenever $x_j=0$.

\subsection{Gradient $\times$ (Input $-$ Baseline)}
\label{app:ga}

\[
\mathrm{GA}_j = (x_j - b_j) \cdot \frac{\partial f}{\partial x_j}(x),
\]
a single-point (non-integrated) gradient attribution using the same
baseline as IG. In the linear arm this coincides exactly with IG (constant
gradient), so we report it mainly for structural parity with the MLP arm,
where GA (single endpoint gradient) and IG (path-integrated gradient
through a nonlinearity) are genuinely different quantities.

\subsection{The Linear-Arm Identity}
\label{app:identity}

With the target matched (Appendix~\ref{app:target-matching}), IG, GA, and
interventional SHAP are \emph{analytically identical} for a linear probe:
all three reduce to $w_j \cdot \mathbb{E}_e[x_{e,j} - \mu_j]$. We verify
this identity numerically (Appendix~\ref{app:supptables},
Table~\ref{tab:app-identity}). We present the linear arm's perfect rank
agreement explicitly as a closed-form sanity check on implementation
correctness, not as an empirical finding that ``SHAP agrees with gradient
methods.'' The substantive cross-method comparison is the MLP arm, where
the ReLU nonlinearity breaks the identity and IG, GA, and SHAP can and do
diverge (Appendix~\ref{app:supptables}, Table~\ref{tab:app-mlp-ranking}).

\clearpage
\section{Causal Evaluation Protocol}
\label{app:causal}

The attribution scores above are correlational, computed from a fixed
model with no intervention. We evaluate causal faithfulness separately, by
directly intervening on SAE feature activations and reading out the
resulting change in \emph{language-model} behavior --- deliberately not
the probe the attributions explain, so that we score faithfulness against
the system the explanation is nominally about, not against the explainer's
own target.

\subsection{Intervention Definitions}
\label{app:interventions}

At \texttt{blocks.7.hook\_resid\_post}, for a chosen feature $i$, we define
three interventions:

\begin{itemize}
  \item \textbf{Ablation}: we encode, set $a_i=0$, and decode, removing the
  feature.
  \item \textbf{Steering}: we encode, set $a_i \leftarrow a_i + \alpha_i$,
  and decode --- additive and signed ($\alpha_i$ may be negative) --- and
  apply it at \emph{every} token position, not only the last token that
  attributions describe (Appendix~\ref{app:limitations}, item 1).
  \item \textbf{Baseline}: we encode and decode with no feature edit (the
  SAE reconstruction baseline), so the measured $\Delta$ is not confounded
  with SAE reconstruction error.
\end{itemize}

\subsection{Readout}
\label{app:readout}

We use the GPT-2 logit difference at the last real token,
$\text{logit}(\text{`` wonderful''}) - \text{logit}(\text{`` awful''})$,
before and after intervention, as the causal readout (both cues resolved
and asserted to be single tokens). This is deliberately \emph{not} the
probe's own prediction, so causal evaluation tests whether a feature
causally matters to the underlying language model's sentiment behavior,
not merely to the probe fit on top of it. We report $\Delta$ both as
$|\Delta|$ (importance) and signed $\Delta$ (direction).

\subsection{Steering-Strength ($\alpha$) Calibration}
\label{app:alpha}

We calibrate a global scalar $\alpha$ via a pilot sweep
($\{0.5,1,2,4\}\times$ the median natural positive-activation scale) over
three top-$|\mathrm{SHAP}|$ pilot features, on 200 \emph{train}-split
examples (never validation or held-out). The readout we use for
calibration differs from the readout we use for evaluation: we choose
$\alpha$ by its effect on a layer-11 residual probe's
$P(\text{positive})$, while the downstream steering scripts read out the
GPT-2 logit difference (Appendix~\ref{app:readout}). This is a deliberate,
acknowledged mismatch: $\alpha$ is a rough ``measurable, unsaturated''
magnitude, not a hyperparameter tuned against the metric we later evaluate
it with.

\paragraph{Selection rule.} We select the smallest $\alpha$ with
$\text{mean}|\Delta P| \geq 0.02$ and saturation fraction (examples with
$P<0.02$ or $P>0.98$) $\leq 0.25$, falling back to the best unsaturated
$\alpha$ if no candidate meets the threshold. The calibration pilot
(Table~\ref{tab:app-alpha-pilot-raw}) recommends $\alpha=0.3316$.

\begin{table}[h]
\centering
\caption{Raw $\alpha$-calibration pilot sweep (probe-readout, 3 pilot features, 200 train examples).}
\label{tab:app-alpha-pilot-raw}
\begin{tabular}{cccc}
\toprule
$\alpha$ & mean$|\Delta P|$ & mean signed $\Delta P$ & saturation frac.\ \\
\midrule
0.3316 & 0.0430 & $-0.0426$ & 0.085 \\
0.6632 & 0.0866 & $-0.0859$ & 0.090 \\
1.3264 & 0.1705 & $-0.1690$ & 0.120 \\
2.6528 & 0.2983 & $-0.2957$ & 0.263 \\
\bottomrule
\end{tabular}
\end{table}

\paragraph{Per-feature $\alpha$-mode.} Given a global $\alpha$, we support
two resolution modes, applied per feature:
\begin{itemize}
  \item \textbf{scaled} (default/primary): $\alpha_i = \alpha \cdot
  \mathrm{scale}_i / \mathrm{median}(\mathrm{scale})$, where
  $\mathrm{scale}_i$ is feature $i$'s mean positive train activation. This
  equalizes the \emph{relative} nudge across features with severalfold
  different natural activation scales.
  \item \textbf{constant} (secondary/robustness check): a flat $\alpha$ for
  every feature. This confounds the $|\text{attribution}|$-vs-$|\Delta|$
  correlation, since $|\Delta|$ then partly reflects how hard a feature was
  pushed relative to its own scale rather than the attribution itself.
\end{itemize}
We apply no decoder-norm correction, since $W_{\text{dec}}$ rows are
approximately unit norm (Appendix~\ref{app:model}).

\subsection{Candidate Selection Strategies}
\label{app:selection}

We evaluate three candidate-selection strategies:

\begin{itemize}
  \item \textbf{random} (headline): we draw $k$ features uniformly from the
  filtered pool. Unbiased.
  \item \textbf{top} (secondary): we take the top-$k$ by $|\mathrm{SHAP}|$,
  then score all four methods on that same set. This range-restricts
  SHAP's own scored variance while leaving the other three methods'
  unrestricted, mechanically deflating SHAP's $\rho$ relative to methods
  that did not drive selection; we flag every \texttt{top}-selection
  faithfulness number with this caveat.
  \item \textbf{all} (higher-power alternative): we score every feature in
  the filtered pool ($n=116$) with no resampling, carrying no seed-draw
  luck. At $n=20$ the confidence interval on Spearman $\rho$ spans most of
  $[-1,1]$, so $\rho\approx0$ from one $k=20$ draw is compatible with both
  a true null and a real moderate effect; we therefore treat this, not the
  $k=20$ random run, as the powered headline global-faithfulness number.
\end{itemize}

\subsection{Faithfulness Metric and Sign Convention}
\label{app:sign-convention}

\begin{itemize}
  \item \textbf{Importance faithfulness}: Spearman $\rho(|\text{attribution
  score}|, |\Delta|)$ --- we ask whether the method's magnitude ranking
  predicts how much the causal effect moves.
  \item \textbf{Directional faithfulness}: Spearman $\rho(\text{signed
  attribution}, \text{oriented }\Delta)$ --- we ask whether the method's
  sign predicts the direction of the causal effect.
  \item \textbf{Sign convention}: steering adds $+\alpha$, so a positive
  attribution should predict a positive $\Delta$. Ablation \emph{removes}
  the feature, so a positive attribution should predict a \emph{negative}
  $\Delta$; we therefore negate raw ablation $\Delta$ before correlating,
  so that ``positive $\rho$'' always means ``faithful'' regardless of mode.
\end{itemize}

\subsection{Multiple-Comparison Correction}
\label{app:bh}

We optionally apply a Benjamini--Hochberg step (family defined per claim
--- e.g.\ the 8 tests of one configuration, or that
$\times\ n_{\text{seeds}}$ for a sweep) on top of the raw faithfulness
tables, as a sidecar that does not alter the original numbers. Our
pipeline's full uncorrected tables report 30+ tests, of which exactly two
clear $p<0.05$ raw --- the expected yield under a complete null. We report
corrected counts, not raw hits, in the main text.

\clearpage
\section{Local (Per-Example) Faithfulness Protocol}
\label{app:local}

Global $\Phi$ is a mean over 2000 examples; averaging can hide per-example
attribution error or manufacture agreement that does not hold
example-by-example. We test this directly with three scripts sharing a
matched CLI surface ($k_{\text{local}}=k_{\text{global}}=20$,
$n_{\text{examples}}=100$) so the two probe arms are comparable.

\subsection{Candidate Construction}
\label{app:local-candidates}

Per sentence $x$, we compute local SHAP $\phi(x)$ with the same explainer
class as the global computation, applied to one row. We construct the
candidate set for that sentence as top-$k$ by $|\phi(x)|$ (local)
$\cup$ top-$k$ by $|\Phi|$ (global), so local and global attribution
compete on equal footing rather than the candidate set structurally
favoring the global method. We keep per-example intervention effects
$\Delta(x,i)$ per example, never averaged across the dataset before
correlating --- this is the entire point of the local analysis, in
contrast to the dataset-mean $\Delta$ used in Appendix~\ref{app:causal}.
Per sentence, we compute Spearman
$\rho(|\text{local }\phi(x)|, |\Delta(x,\cdot)|)$ and
$\rho(|\text{global }\Phi|, |\Delta(x,\cdot)|)$ on the same candidate set,
and summarize them as the mean $\rho$ across sentences, plus the fraction
of sentences on which local beats global.

\subsection{The Activity Confound and Its Control}
\label{app:activity-confound}

Globally-top features (by $|\Phi|$) are usually \emph{inactive} in any
single sentence. Ablating an inactive feature moves the logit by exactly 0
(there is nothing to remove), while local $|\phi(x)|$ tracks per-sentence
activity by construction. A chunk of any apparent ``local SHAP beats global
$\Phi$'' margin is therefore just ``local knows which features are on in
this sentence,'' not evidence about ranking quality among features that
matter.

\paragraph{Block A / Block B.} Block A is the full candidate set described
above. Block B restricts candidates to those the sentence actually
activates ($\text{activation}>0$) before re-scoring. A local win that
survives Block B is evidence about ranking quality among \emph{active}
features, not merely about knowing which features fired. We take Block B,
not Block A, as the headline local-vs-global comparison reported in the
main text, since Block A is dominated by the inactive tie block. 

\clearpage
\section{Multi-Feature (Group) Steering Protocol}
\label{app:group}

In a separate and secondary, larger-scale causal experiment, we build a single steering
\emph{direction} (not a single-feature intervention) from the top-$k$
features of each attribution method and sweep effect against cost.

\begin{itemize}
  \item \textbf{Vector construction}: for method $m$, we take the top-$k$
  features by $|\text{score}|$ from the shared 116-feature filtered pool
  (the same pool for both arms, so top-$k$ at equal $k$ draws from the
  same candidate space), then set
  $v = \sum_i \text{score}_i \cdot W_{\text{dec}}[i]$, unit-normalized.
  \item \texttt{actdiff} \textbf{baseline}: a no-attribution-math baseline
  --- top-$k$ by $|\overline{a}^{+} - \overline{a}^{-}|$ (mean positive
  minus mean negative activation), weighted by the signed difference
  itself. We score it once (arm-independent, since it uses no probe) and
  fan it out to both arms' plots, tagged \texttt{arm\_independent: true}.
  \item \textbf{No MLP \texttt{probe} curve, by construction}: the MLP's
  \texttt{probe\_saliency} ($\|W_1[i,:]\|_2$) is unsigned, and our
  vector-construction formula uses score as a \emph{signed} weight; summing
  unsigned magnitudes with the decoder would add positive- and
  negative-sentiment directions with the same sign, producing a direction
  that means nothing. We retain the logistic arm's \texttt{probe} curve
  because \texttt{coef\_[0]} is signed.
  \item \textbf{Sweep}: we sweep over $k$ and $\alpha$ grids (full grid and
  row counts in Appendix~\ref{app:group-full}).
  \item \textbf{Effect} = mean $\Delta$ logit-diff (wonderful $-$ awful);
  \textbf{cost} = mean $\Delta$ perplexity (steered vs.\ clean baseline).
  \item \textbf{Baseline differs from Appendix~\ref{app:causal}}: this
  experiment never touches the SAE at inference --- we add
  $\alpha \cdot v$ directly to the clean residual stream, with no
  encode/decode --- so its baseline is a \emph{clean} forward pass, whereas
  the per-feature steering/ablation experiments baseline through an
  \emph{encode$\to$decode reconstruction}. Both are internally consistent,
  but effect sizes across the two families are on different scales; we
  never compare them numerically, only the \emph{shapes} of the
  effect-vs-cost curves.
\end{itemize}

\begin{figure}[H]
  \centering
  \begin{subfigure}[b]{0.48\linewidth}
    \includegraphics[width=\linewidth]{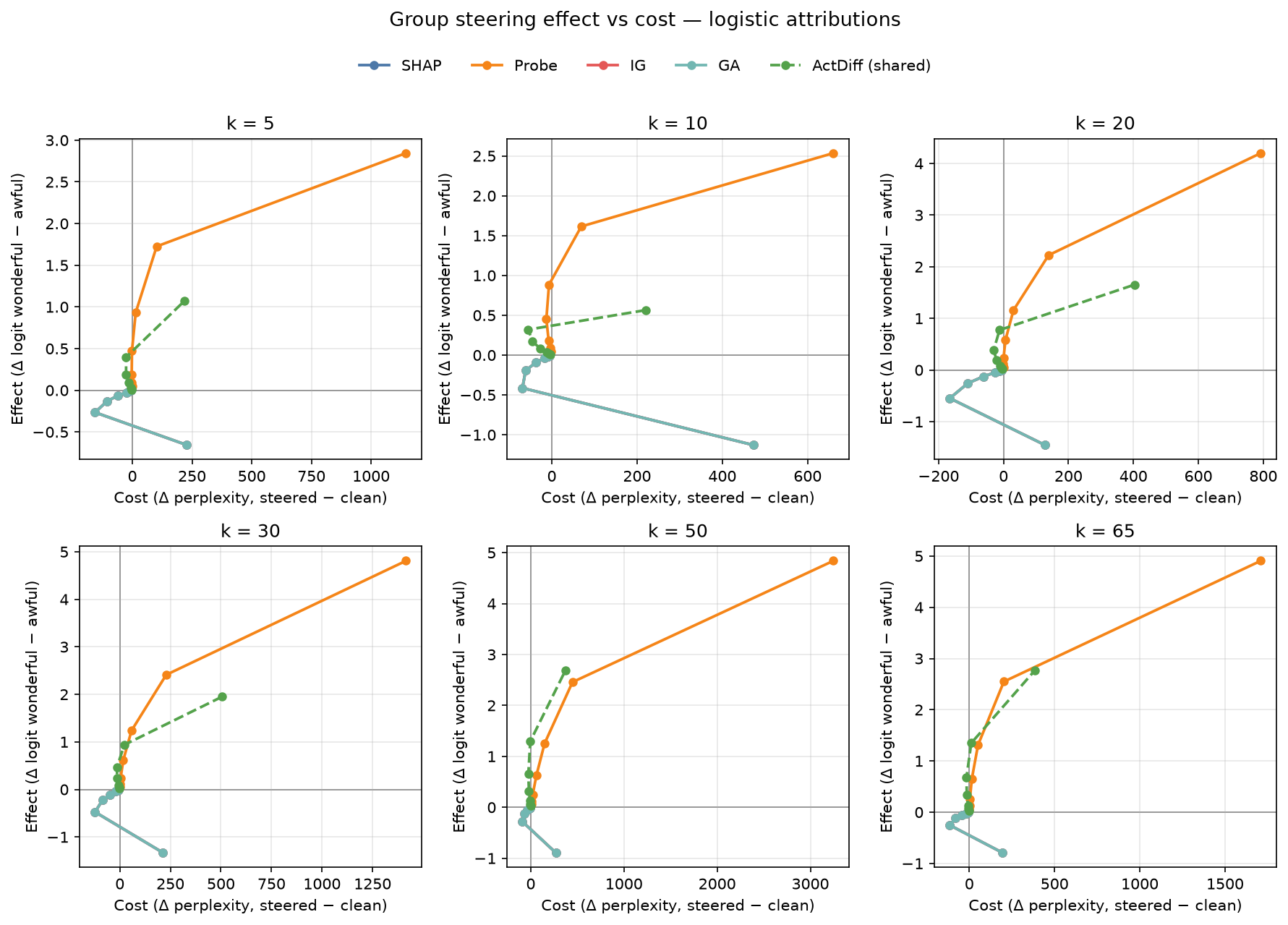}
    \caption{Logistic arm.}
  \end{subfigure}
  \hfill
  \begin{subfigure}[b]{0.48\linewidth}
    \includegraphics[width=\linewidth]{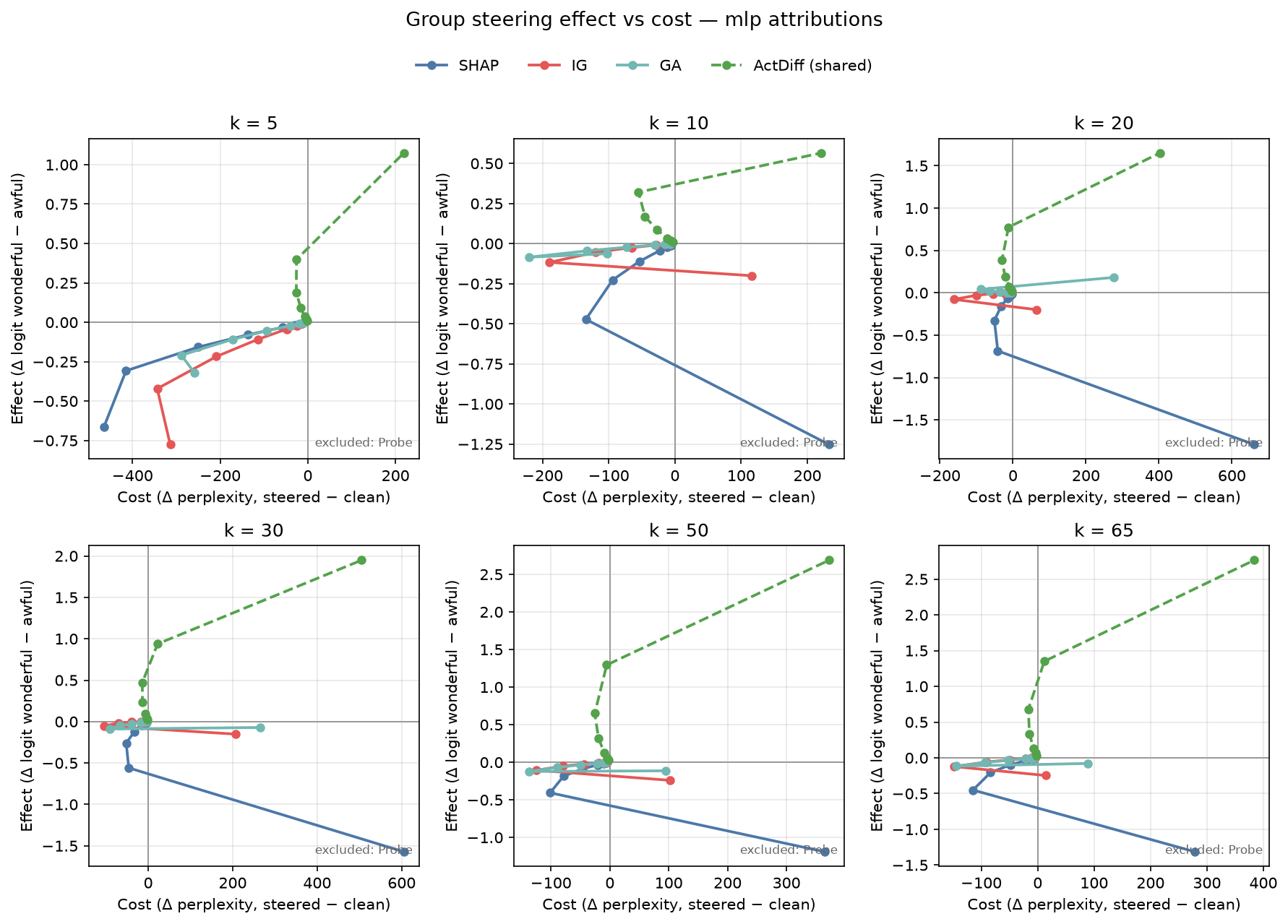}
    \caption{MLP arm.}
  \end{subfigure}
  \caption{Full group-steering effect-vs-cost grid ($k\times\alpha$), one panel per method per arm. This experiment's baseline is a clean forward pass (no SAE encode/decode), unlike the per-feature steering/ablation experiments' encode$\to$decode-reconstruction baseline (Appendix~\ref{app:group}); we compare curve \emph{shapes} across arms, never magnitudes.}
  \label{fig:app-group-steering-grid}
\end{figure}

We show full parameter grids for both arms in
Figure~\ref{fig:app-group-steering-grid}; the main text reports a $k=20$
spot-check drawn from this same sweep.

\clearpage
\section{Known-Ground-Truth Validation Protocol (Minimal Pairs)}
\label{app:minimal-pairs}

Every faithfulness number we compute on natural SST-2 text has an unknown
ground-truth causal importance for any given SAE feature, so a near-zero
$\rho$ has two incompatible readings that the main pipeline alone cannot
separate: (a) the attribution methods really are unfaithful, or (b) the
harness itself ($k=20$, $\alpha$ calibrated on a different readout,
last-token attribution vs.\ all-token intervention) cannot detect
faithfulness even when it is present. We build stimuli where we know a real
sentiment change by construction and ask whether the harness recovers it.

\subsection{Stimulus Construction}
\begin{itemize}
  \item \textbf{\texttt{remove} (primary)}: we take held-out sentences
  containing \emph{exactly one} lexicon word (from a hand-curated
  positive/negative sentiment word list) and pair each with that single
  token deleted; everything else is byte-identical. We require ``exactly
  one,'' not ``at least one,'' because with two sentiment words the
  remaining one keeps carrying sentiment and the pair is no longer a
  single-cause edit.
  \item \textbf{\texttt{insert} (secondary)}: we append ``\,it was
  \textlangle word\textrangle\,.'' to lexicon-free sentences. This is less
  minimal (it adds a clause) but tests the insertion direction that
  \texttt{remove} pairs cannot.
\end{itemize}
We know the expected direction: removing a positive word should lower the
logit difference; adding one should raise it.

\subsection{Three Measurements}
\begin{enumerate}
  \item \textbf{Readout validity} is the sign accuracy of the text-level
  $\Delta$ against the known direction. This is a precondition for every
  other faithfulness claim in the paper: a readout that cannot detect a
  deliberately inserted sentiment word cannot support any faithfulness
  claim.
  \item \textbf{Feature recovery} (the actual ground truth) are features
  whose activation moved between the pair demonstrably carry the edit. We
  label those positive and distractors negative, giving a binary ground
  truth with a known chance level (AUROC $0.5$), and score it per method.
  \item For \textbf{Causal recovery}, we ablate each candidate in the
  edited sentence, measure how much of the edit's effect it was supplying,
  and correlate that continuous ground-truth causal score against each
  attribution.
\end{enumerate}

\clearpage
\section{Statistical Reporting Convention: Scored Support Only}
\label{app:stats-conventions}

We restrict all correlations in this paper to the non-zero/scored feature
support, never computing them over the full 32{,}768-feature space. The
latter would report agreement about which features were \emph{excluded}
rather than agreement about importance among included features.

\clearpage
\section{Supplementary Tables and Figures}
\label{app:supptables}

\subsection{Linear-Arm Identity Check and MLP-Arm Exact $p$-Values}

\begin{quote}
\textit{Closed-form identity check (linear probe, margin target):}
$\max|\mathrm{IG}-\Phi| = 6.32\times10^{-7}$ (relative $5.06\times10^{-6}$),
$\max|\mathrm{GA}-\Phi| = 6.32\times10^{-7}$. IG, GA, and
interventional SHAP coincide analytically for a linear probe; we obtain
pairwise $\rho = 1.000$ ($n=1301$ non-zero-coefficient features on this
fit) as an identity, not an empirical result.
\end{quote}

\begin{table}[h]
\centering
\caption{MLP-arm attribution agreement (Spearman $\rho$ on $|\text{score}|$, $n=116$).}
\label{tab:app-mlp-ranking}
\begin{tabular}{lcc}
\toprule
Pair & $\rho$ & $p$ \\
\midrule
Probe vs.\ IG   & 0.024 & 0.800 \\
Probe vs.\ SHAP & 0.004 & 0.965 \\
Probe vs.\ GA   & 0.037 & 0.696 \\
IG vs.\ SHAP    & 0.617 & $<0.0001$ \\
IG vs.\ GA      & 0.804 & $<0.0001$ \\
SHAP vs.\ GA    & 0.418 & $<0.0001$ \\
\bottomrule
\end{tabular}
\end{table}
\label{tab:app-identity}

\subsection{Selection-Bias and Sign-Stratified Checks}
\label{app:selection-table}

\begin{table}[h]
\centering
\caption{SHAP faithfulness across candidate-selection strategies (steering), $k=20$ except \texttt{all} ($k=116$). \texttt{top} biases SHAP's $\rho$ downward relative to other methods (Appendix~\ref{app:selection}); \texttt{top-pos}/\texttt{top-neg} test the sign-inversion hypothesis without mixed-sign cancellation.}
\label{tab:app-selection-variants}
\begin{tabular}{lccccc}
\toprule
Arm : kind & random & top & top-pos & top-neg & all ($k$=116) \\
\midrule
Linear : imp & $-0.083$ & $+0.356$ & $+0.158$ & $+0.167$ & $+0.241$ \\
Linear : dir & $+0.119$ & $-0.215$ & $-0.310$ & $+0.075$ & $+0.034$ \\
MLP : imp    & $-0.298$ & $+0.471$ & $+0.771^{**}$ & $+0.195$ & $+0.210$ \\
MLP : dir    & $+0.003$ & $-0.220$ & $-0.383$ & $+0.189$ & $-0.005$ \\
\bottomrule
\end{tabular}
\vspace{2pt}
{\footnotesize $^{**}p=0.0001$; all other cells $p>0.09$, uncorrected.}
\end{table}

\subsection{Alpha-Mode Robustness (Scaled vs.\ Constant)}
\label{app:alpha-mode-table}

\begin{table}[h]
\centering
\caption{Alpha-mode sensitivity: scaled vs.\ constant per-feature $\alpha$ (\texttt{steering}, \texttt{random}, $k=20$).}
\label{tab:app-alpha-mode}
\begin{tabular}{llcc}
\toprule
Arm & Method & \rhoimp\ (scaled / constant) & \rhodir\ (scaled / constant) \\
\midrule
Linear & SHAP/IG/GA & $-0.083$ / $-0.114$ & $+0.119$ / $+0.202$ \\
Linear & Probe      & $-0.186$ / $-0.081$ & $-0.011$ / $-0.089$ \\
MLP    & SHAP       & $-0.298$ / $-0.377$ & $+0.003$ / $+0.081$ \\
MLP    & Probe saliency & $-0.146$ / $+0.003$ & n/a \\
\bottomrule
\end{tabular}
\end{table}

A large gap here would indicate that constant-$\alpha$ steering numbers
substantially reflect activation scale rather than faithfulness; we report
both modes regardless of which way the comparison comes out
(Appendix~\ref{app:limitations}, item 9).

\subsection{Multi-Seed Replication}
\label{app:seed-sweep}

We pool ten seeds' candidate draws into one intervention pass and slice
them back out per seed (\texttt{steering}, \texttt{random},
\texttt{scaled} $\alpha$, $k=20$), so replication costs a fraction of ten
independent reruns.

\begin{table}[h]
\centering
\caption{Multi-seed faithfulness sweep (10 seeds, $k=20$, scaled $\alpha$). ``Ref $p$'' is the fraction of seeds with $|\rho|$ at least the published seed-0 draw's.}
\label{tab:app-seed-sweep}
\begin{tabular}{llccccc}
\toprule
Arm & Test & Mean $\rho$ & Std & 95\% CI & Wilcoxon $p$ & Ref $p$ (seed 0) \\
\midrule
Linear & SHAP/IG/GA : imp & $+0.264$ & 0.184 & [0.149, 0.382] & 0.0039 & 1.000 \\
Linear & SHAP/IG/GA : dir & $-0.022$ & 0.227 & [$-$0.155, 0.125] & 0.770 & 0.700 \\
Linear & Probe : imp & $+0.140$ & 0.197 & [0.021, 0.262] & 0.0488 & 0.500 \\
Linear & Probe : dir & $+0.320$ & 0.188 & [0.203, 0.433] & 0.0039 & 1.000 \\
\midrule
MLP & SHAP : imp & $+0.189$ & 0.254 & [0.024, 0.342] & 0.0625 & 0.500 \\
MLP & SHAP : dir & $-0.064$ & 0.140 & [$-$0.154, 0.020] & 0.322 & 1.000 \\
MLP & GA : imp & $+0.180$ & 0.207 & [0.049, 0.308] & 0.0371 & 0.600 \\
\bottomrule
\end{tabular}
\vspace{2pt}
{\footnotesize Pooled BH: 0/80 tests survive (linear), 0/70 (MLP), at FDR 0.05.}
\end{table}

\subsection{Underpowered $k=20$ causal validation results}
\label{app:k-20}

\begin{table}[H]
\centering
\caption{Underpowered causal faithfulness ($k=20$, steering, random selection, $n_\text{eval}=1000$). For the linear arm, IG and GA are analytically identical to SHAP and omitted as separate rows.}
\label{tab:headline-faithfulness}
\begin{tabular}{llcccc}
\toprule
Arm & Mode & Method & $\rho_\text{imp}$ ($p$) & $\rho_\text{dir}$ ($p$) \\
\midrule
Linear & Steering & SHAP/IG/GA & $-0.083$ (0.729) & $+0.119$ (0.618) \\
Linear & Steering & Probe       & $-0.186$ (0.431) & $-0.011$ (0.965) \\
Linear & Ablation & SHAP/IG/GA & $-0.045$ (0.850) & $-0.164$ (0.490) \\
Linear & Ablation & Probe       & $-0.183$ (0.439) & $+0.448$ (0.048) \\
\midrule
MLP & Steering & SHAP           & $-0.298$ (0.202) & $+0.003$ (0.990) \\
MLP & Steering & Probe saliency & $-0.146$ (0.540) & n/a \\
MLP & Steering & IG             & $-0.227$ (0.336) & $-0.059$ (0.806) \\
MLP & Steering & GA             & $+0.202$ (0.394) & $+0.171$ (0.470) \\
MLP & Ablation & SHAP           & $+0.194$ (0.413) & $-0.140$ (0.557) \\
MLP & Ablation & Probe saliency & $-0.403$ (0.078) & n/a \\
MLP & Ablation & IG             & $+0.114$ (0.631) & $-0.290$ (0.215) \\
MLP & Ablation & GA             & $-0.068$ (0.777) & $-0.197$ (0.405) \\
\bottomrule
\end{tabular}
\end{table}

\paragraph{Continuous (partial-Spearman) activity control.} We compute a
partial Spearman of $|\text{attribution}|$ vs.\ $|\Delta|$ controlling
continuously for $|\text{activation}|$ (rather than the binary
active/inactive split of Block B), keeping the whole candidate set: local
SHAP retains partial $\rho = +0.184$ (linear) / $+0.150$ (MLP).

\subsection{Group-Steering Full Sweep}
\label{app:group-full}

This is the full parameter grid underlying the main text's $k=20$
spot-check: $k \in \{5,10,20,30,50,65\} \times \alpha \in
\{0.5,1,2,5,10,20,50\}$, giving 7 vectors (4 logistic + 3 MLP) $\times$ 6
$k$ $\times$ 7 $\alpha$ = 294 arm-specific rows; we also score
\texttt{actdiff} once at $6\times7=42$ and duplicate it across both arms
(84 rows, \texttt{arm\_independent: true}) --- 378 total rows from 336
forward passes.

\subsection{Lexical Simplicity Control}
\label{app:lexical}

We use the full 10{,}103-row held-out split as the population here, not
the 2000-row canonical eval sample used everywhere else in the paper.

\begin{table}[h]
\centering
\caption{Lexical simplicity control (full held-out split, $n=10{,}103$).}
\label{tab:app-lexical-simplicity}
\begin{tabular}{lc}
\toprule
Quantity & Value \\
\midrule
Mean active filtered features / example (of 116) & 23.06 \\
Mean pairwise Jaccard, top-20 $|\Phi|$ features & 0.211 \\
Mean $|r(\text{activation}, \text{has-lexicon-word})|$, top-20 & 0.063 \\
Lexicon bag-of-words accuracy / AUROC & 0.629 / 0.676 \\
SAE probe accuracy / AUROC (held-out) & 0.761 / 0.838 \\
Accuracy gap (SAE probe $-$ lexicon) & $+0.132$ \\
\bottomrule
\end{tabular}
\end{table}

By the reading thresholds we fixed in advance (mean $|r| \geq 0.25$ for
``strong lexical tracking''; within 0.03 accuracy for
``lexicon-near-probe''): the observed mean $|r|=0.063$ is well below the
strong-tracking threshold, and the 0.132 accuracy gap is well above the
near-tie threshold. We take this as indicating that the top SAE features
are not simply lexical sentiment-word detectors, and that non-lexical
signal remains in the representation for attribution methods to
potentially explain.

\subsection{All-token global faithfulness}
\label{app:all-token-global-faithfulness}
\begin{table}[h!]
\centering
\caption{Powered global faithfulness replication ($k=116$, steering, all filtered features, BH-corrected within arm).}
\label{tab:all-selection-faithfulness}
\begin{tabular}{llcc}
\toprule
Arm & Method & $\rho_\text{imp}$ ($q$) & $\rho_\text{dir}$ ($q$) \\
\midrule
Linear & SHAP/IG/GA & $+0.241$ ($q$=0.019)* & $+0.034$ ($q$=0.719) \\
Linear & Probe      & $+0.134$ ($q$=0.241)  & $+0.356$ ($q$=0.0007)* \\
\midrule
MLP & SHAP           & $+0.210$ ($q$=0.126) & $-0.005$ ($q$=0.958) \\
MLP & Probe saliency & $+0.018$ ($q$=0.958) & n/a \\
MLP & IG             & $+0.152$ ($q$=0.242) & $+0.019$ ($q$=0.958) \\
MLP & GA             & $+0.195$ ($q$=0.126) & $+0.030$ ($q$=0.958) \\
\bottomrule
\multicolumn{4}{l}{\footnotesize *survives FDR 0.05. Linear arm: 4/8 tests survive; MLP arm: 0/7.}
\end{tabular}
\end{table}

\clearpage
\section{Known Limitations}
\label{app:limitations}

\begin{enumerate}
  \item \textbf{Attribution is last-token; intervention is all-token.}
  Attributions describe the SAE vector at one sequence position, while the
  intervention edits every token position. Part of any faithfulness gap is
  this scope mismatch, not necessarily attribution error.
  \item \textbf{We calibrate $\alpha$ against a different readout than we
  evaluate it with} (Appendix~\ref{app:alpha}). We use a layer-11 residual
  probe's $P(\text{positive})$ at calibration time, and a GPT-2 logit
  difference at evaluation time.
  \item \textbf{The MLP arm's \texttt{probe\_saliency} is magnitude-only}
  ($\|W_1[i,:]\|_2$, no sign). This is valid for importance/ranking comparisons,
  invalid for any directional or causal-sign comparison. We exclude it
  from group-steering vector construction for the same reason.
  \item \textbf{\texttt{selection=top} biases SHAP's faithfulness $\rho$
  downward} relative to the other three methods, by range-restricting only
  SHAP's scored variance (Appendix~\ref{app:selection}). We never present
  a \texttt{top} number without this caveat.
  \item \textbf{Faithfulness $\rho$ are small-$n$}, with no
  multiple-comparison control by default (Appendix~\ref{app:bh}).
  \item \textbf{The MLP probe's reported validation accuracy is mildly
  optimistic} --- we select its inputs using a val-derived (label-free)
  frequency filter.
  \item \textbf{The linear arm's rank-agreement $\rho=1.000$ is a
  closed-form identity} (Appendix~\ref{app:identity}), not an empirical
  finding --- it is not evidence that SHAP and gradient methods agree in
  general.
\end{enumerate}

\end{document}